\documentclass{article} 
\usepackage[final]{colm2025_conference}

\usepackage{hyperref}
\usepackage{placeins}
\usepackage[utf8]{inputenc}  
\usepackage{microtype}
\usepackage{hyperref}
\usepackage{url}
\usepackage{float} 
\usepackage{amsmath}
\usepackage{graphics}

\usepackage{booktabs}
\usepackage{tabularx}
\usepackage{makecell}
\usepackage{adjustbox}
\usepackage{multirow}
\usepackage{caption} 
\usepackage{subcaption} 
\usepackage[most]{tcolorbox}
\usepackage{enumitem}
\usepackage{fontawesome5}
\newtcolorbox{promptbox}{breakable, colback=white, colframe=black,
  boxrule=0.6pt, arc=1mm, left=1.2ex, right=1.2ex, top=1ex, bottom=1ex}
\setlist[itemize]{leftmargin=*, itemsep=2pt, topsep=2pt}

\newcolumntype{Y}{>{\centering\arraybackslash}X}

\DeclareUnicodeCharacter{2212}{-} 

\definecolor{darkblue}{rgb}{0, 0, 0.5}
\hypersetup{colorlinks=true, citecolor=darkblue, linkcolor=darkblue, urlcolor=darkblue}

\title{Leveraging Author-Specific Context for Scientific Figure \\ 
Caption Generation: 3rd SciCap Challenge }

\author{%
Watcharapong Timklaypachara$^{1,2}$ \,
Monrada Chiewhawan$^{1,2}$ \,
Nopporn Lekuthai$^{1,2}$ \\
\textbf{Titipat Achakulvisut$^{1}$} \\
$^1$Department of Biomedical Engineering, Faculty of Engineering, Mahidol University \\
$^2$Faculty of Medicine Ramathibodi Hospital, Mahidol University, Bangkok, Thailand
}

\begin{document}

\maketitle

\centerline{\faGithub\ \quad \href{github.com/biodatlab/scicap-titipapa}{\texttt{https://github.com/biodatlab/scicap-titipapa}}}
\vspace{1em}

\begin{abstract}

Scientific figure captions require both accuracy and stylistic consistency to convey visual information. Here, we present a domain-specific caption generation system for the 3rd SciCap Challenge that integrates figure-related textual context with author-specific writing styles using the LaMP-Cap dataset. Our approach uses a two-stage pipeline: Stage 1 combines context filtering, category-specific prompt optimization via DSPy's MIPROv2 and SIMBA, and caption candidate selection; Stage 2 applies few-shot prompting with profile figures for stylistic refinement. Our experiments demonstrate that category-specific prompts outperform both zero-shot and general optimized approaches, improving ROUGE-1 recall by +8.3\% while limiting precision loss to −2.8\% and BLEU-4 reduction to −10.9\%. Profile-informed stylistic refinement gains of 40–48\% in BLEU scores and 25–27\% in ROUGE. Our system demonstrates that combining contextual understanding with author-specific stylistic adaptation can generate captions that are scientifically accurate and stylistically faithful to the source paper.
\end{abstract}
\section{Introduction}
Scientific figure captions play a crucial role in helping diverse audiences interpret and understand complex visual information \citep{Lundgard2021}. However, manual caption writing is time-consuming and often inconsistent, particularly when authors must generate numerous captions while maintaining stylistic coherence throughout their manuscripts. Automatic caption generation provides a promising solution to enhance scientific communication efficiency and reduce the burden on researchers \citep{Hsu2021}. Recent advances in large language models (LLMs) have demonstrated the ability to comprehend complex scientific content and adapt to distinctive linguistic patterns \citep{Zanotto2024}. This capability enables the development of context-aware systems that can generate specific author writing styles, potentially reducing manual editing.  
Our approach uses the LaMP-Cap dataset with contextual information to generate context-grounded and personalized scientific captions. Here, we present a two-stage scientific figure caption generation at the 3rd SciCap Challenge. The first stage involves context filtering,  prompt optimization to generate category-focused captions, and candidate selection. The second stage incorporates relevant profile information to improve stylistic elements consistent with the source paper. We found that combining these approaches improves the alignment of the generated caption.

\begin{figure}
    \centering
    \includegraphics[width=0.6\linewidth]{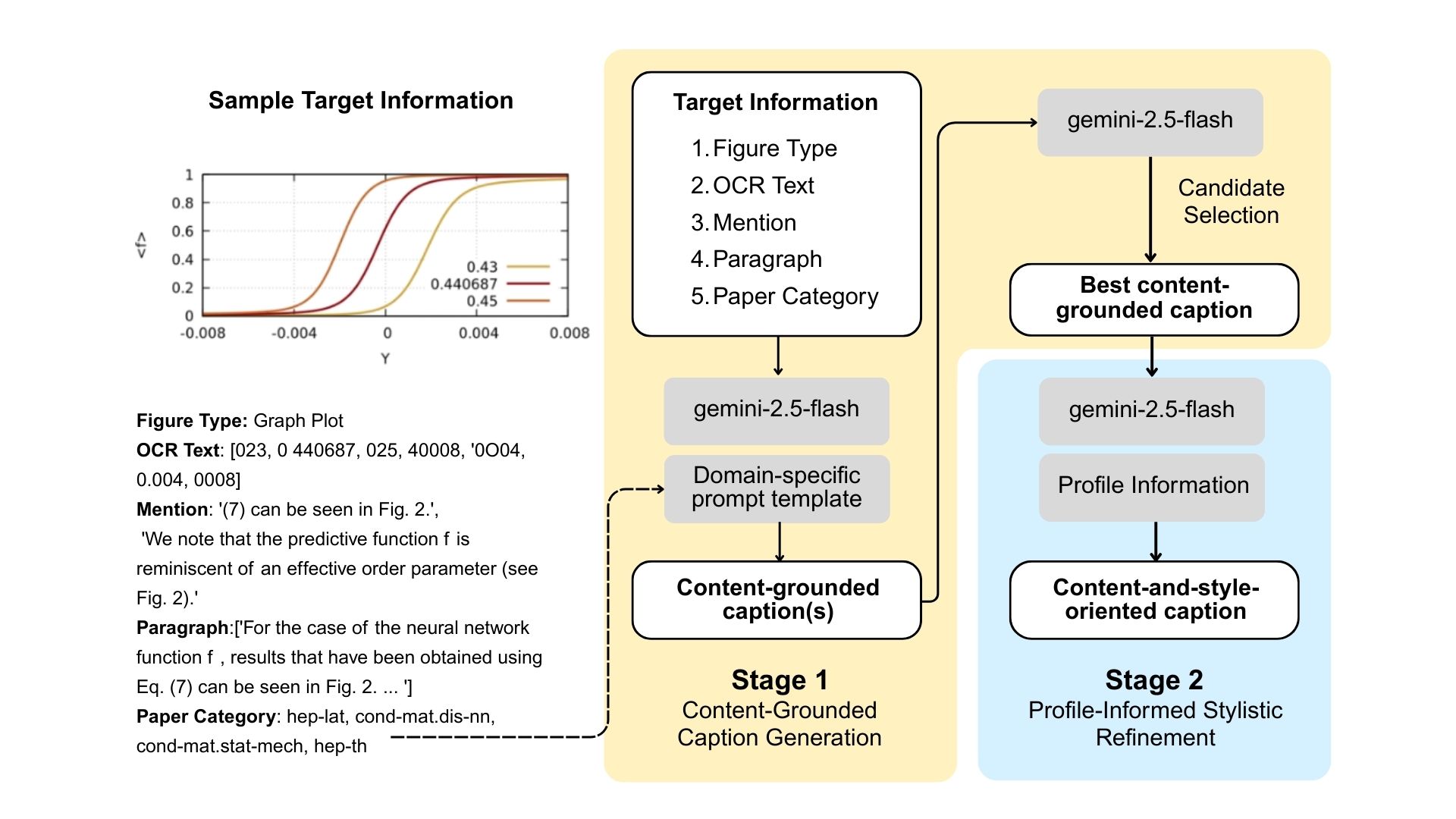}
    \caption{Overview of our multi-stage caption generation framework}
    \label{fig:overall}
\end{figure}

\textbf{\section{Data Processing}} 
We used the LaMP-CAP dataset for caption generation. The dataset consists of 110,828 scientific articles with 80:10:10 train/test/validation splits. Each article includes one target figure and up to three associated profile figures. Each figure contains mentioned text, accompanying paragraph, OCR texts, caption length, and figure type as a context for the caption generation. During prompt optimization, we exclusively used the training set to generate instruction templates per paper category. This dataset contains 8 fields including Computer Science, Economics, Electrical Engineering and Systems Science, Mathematics, Physics, Quantitative Biology, Quantitative Finance, and Statistics. There are 155 unique categories with 50,110 single category papers and 60,718 papers with multiple categories (1.83 ± 0.95 categories per paper on average).
\textbf{\section{Methodology}} 
Zero-shot prompting with paragraph information produces noisy captions with irrelevant content and hallucinations. We developed a two-stage pipeline to address this issue (Figure~\ref{fig:overall}). Stage 1 generates content-grounded captions while minimizing irrelevant content. Stage 2 refines captions using profile-informed stylistic adjustments for factual accuracy and author alignment.

\subsection{\textbf{Stage 1: Content-Grounded Caption Generation} }
\subsubsection{Filtering Irrelevant Information}
We adapted sentence-based filtering from \citet{Li2024} using Flan T5 as a relevance scorer. Input paragraphs are segmented into chunks using spaCy, and conditional log-likelihood between chunks and mentioning text is estimated. Chunks are retained if they exceed baseline null-prompt likelihood by threshold $\lambda$ = 1.2. This filter is applied only to the training set to obtain category-specific prompt templates. We try both filtering and non-filtering approaches in our experiments to evaluate the impact on caption quality. 
\subsubsection{Category-level Prompt Optimization }
Scientific papers within domains share similar terminology and structural patterns, but papers often span multiple categories (e.g., Mathematics, Computer Science, Economics). We develop category-focused prompt templates using MIPROv2 and SIMBA from DSPy Toolkit \citep{OpsahlOng2024} \citep{Vach2025}. MIPROv2 creates instruction-example pairs from bootstrap templates, optimizing for ROUGE-L precision via Bayesian surrogate modeling. SIMBA then applies feedback-driven optimization to generate reasoning paths and detect problematic cases, incorporating improvement rules into the updated configurations. Papers with multiple categories proceed to the caption candidate selection stage, while single-category papers use their domain-specific caption directly.
\subsubsection{Caption Candidate Selection}
One caption candidate is generated per paper category using its corresponding category-focused prompt template. For papers with multiple categories, we used an LLM to rank and output the best caption candidate. Using Gemini-2.5 Flash, the LLM was given (i) the target figure information without the ground-truth caption, (ii) the complete list of category-wise candidates for that paper, and (iii) strict instructions (See \hyperref[fig:caption-ranking-side]{Appendix}) to minimize hallucination and select a single optimal caption to represent the paper. For papers with only one category, the caption was taken directly from the optimized prompt for that category.

 \subsection{\textbf{Stage 2: Profile-Informed Stylistic Refinement} }

We hypothesized that content-grounded captions lacked personalization and precision due to irrelevant content. To address this, we applied few-shot prompting with two goals: (i) enforcing a ±15\% caption length limit for conciseness, and (ii) personalizing captions using author-specific stylistic patterns from profile figures. Profile captions served as structural references under the assumption of consistent writing style within each paper.

 \section{\textbf{Evaluation} }

 We evaluate caption quality using BLEU-1 to BLEU-4 and ROUGE-1, ROUGE-2, and ROUGE-L metrics to evaluate how closely the generated captions match corresponding reference captions in terms of lexical overlap and structural similarity. The ROUGE scores yield precision, recall, and F1-score values to evaluate over- and under-generation tendencies.

\subsection{Result of Development Set}

\subsubsection{\textbf{Effect of Sentence-Level Filtering on Prompt Optimization} }

We compared the impact of sentence-level filtering on paragraph data with the effects of prompt optimization to evaluate their respective contributions. We examined prompt optimization under two conditions: (a) using the original target figure’s data, and (b) replacing the original paragraph with its filtered version. The result (shown in Table \ref{tab:prompt-opt}) indicated that the filtered paragraph setting (b) improved all ROUGE score recalls: ROUGE-1 improved by 11.8\% (0.4867 to 0.5443), ROUGE-2 by 15.7\% (0.2223 to 0.2572), and ROUGE-L by (0.4068 to 0.4430). However, we observe modest declines in BLEU (5.1 to 7.7\% decrease) and ROUGE precision (13.4 to 17.6\% decrease). This trade-off indicates that filtering during prompt optimization improves recall metrics while reducing precision, suggesting a shift toward more comprehensive but less precise outputs.

\subsubsection{\textbf{Prompt Performance } }

We evaluated three prompting strategies under setting (b): (i) Zero Shot i.e., using only target information without any profile information, (ii) Non-category-focused (MIPROv2 and SIMBA), and (iii) Category-focused (MIPROv2 and SIMBA). As shown in Table \ref{tab:prompt-opt}, Category-focused achieved the best balance, keeping ROUGE-1 and ROUGE-L precision within −2.8\% and −2.9\% of Zero Shot while limiting BLEU-4 loss to −10.9\% (vs. −27.3\% for Non-category-focused). It improved recall by +6.4\% to +9.6\% over (i) and limited F-measure drops to ~4\%. In contrast, Non-category-focused produced inflated recall gains (+26–32\%) but severe precision drops (−35–39\%), reflecting verbose and off-topic content. These results support Category-focused prompting as a more controlled way to enhance recall without sacrificing precision.

 \subsection{Result of Test Set}

 \subsubsection{Caption Candidate Selection Result}
We evaluated caption quality using Gemini-2.5-flash as a reranker, comparing the selected captions against the pool of all generated captions. The selected captions achieved higher alignment performance (Table \ref{tab:caption-performance-sub}). BLEU scores showed marginal improvements across all n-gram levels (+0.13\% to +1.56\%). ROUGE precision exhibited more consistent gains (+3.46\% to +4.43\%), and ROUGE F-measure improved modestly (+0.61\% to +1.51\%). In contrast, ROUGE recall decreased slightly (−1.75\% to −2.37\%), suggesting that the reranker prioritizes precision and stylistic consistency at the expense of broader content coverage.

\begin{table}[t]
\centering
\scriptsize
\renewcommand{\arraystretch}{0.95}

\begin{subtable}{\linewidth}
\centering
\subcaption*{\textbf{(a) BLEU \& ROUGE-1 results}}
\begin{adjustbox}{max width=0.9\textwidth}
\begin{tabular}{lccccccc}
\toprule
\textbf{Condition} & B1 & B2 & B3 & B4 & R1-P & R1-R & R1-F \\
\midrule
Prompt Opt.\ -- Orig.\ Para.\ (a) & 0.2706 & 0.1721 & 0.1178 & 0.0840 & 0.3800 & 0.4867 & 0.3682 \\
Prompt Opt.\ -- Filt.\ Para.\ (b) & \makecell{0.2498 \\ (-7.7\%)} & \makecell{0.1614 \\ (-6.2\%)} 
& \makecell{0.1110 \\ (-5.8\%)} & \makecell{0.0805 \\ (-5.1\%)} 
& \makecell{0.3296 \\ (-13.4\%)} & \makecell{0.5443\textbf{ }\\ (+11.8\%)} 
& \makecell{0.3529 \\ (-4.2\%)} \\
\midrule
Zero Shot (i)       & \textbf{0.2998} & \textbf{0.2004} & \textbf{0.1436} & \textbf{0.1088} & \textbf{0.4151} & 0.4599 & \textbf{0.3937} \\
Non-cat.\ Prompt (ii) & \makecell{0.2365 \\ (-21.1\%)} & \makecell{0.1547 \\ (-22.8\%)} 
& \makecell{0.1079 \\ (-24.9\%)} & \makecell{0.0791 \\ (-27.3\%)} 
& \makecell{0.2665 \\ (-35.8\%)} & \makecell{\textbf{0.6089} \\ (+32.4\%)} 
& \makecell{0.3260 \\ (-17.2\%)} \\
Cat.-focused Prompt (iii) & \makecell{0.2699 \\ (-10.0\%)} & \makecell{0.1819 \\ (-9.3\%)} 
& \makecell{0.1293 \\ (-10.0\%)} & \makecell{0.0970 \\ (-10.9\%)} 
& \makecell{0.4034 \\ (-2.8\%)}  & \makecell{0.4979 \\ (+8.3\%)} 
& \makecell{0.3774 \\ (-4.1\%)} \\
\bottomrule
\end{tabular}
\end{adjustbox}
\end{subtable}

\begin{subtable}{\linewidth}
\centering
\subcaption*{\textbf{(b) ROUGE-2 \& ROUGE-L results}}
\begin{adjustbox}{max width=0.9\textwidth}
\begin{tabular}{lcccccc}
\toprule
\textbf{Condition} & R2-P & R2-R & R2-F & RL-P & RL-R & RL-F \\
\midrule
Prompt Opt.\ -- Orig.\ Para.\ (a) & 0.1800 & 0.2223 & 0.1740 & 0.3145 & 0.4068 & 0.3032 \\
Prompt Opt.\ -- Filt.\ Para.\ (b) & \makecell{0.1522 \\ (-15.4\%)} & \makecell{0.2572 \\ (+15.7\%)} 
& \makecell{0.1636 \\ (-6.0\%)} & \makecell{0.2591 \\ (-17.6\%)} 
& \makecell{0.4430 \\ (+8.9\%)}  & \makecell{0.2790 \\ (-8.0\%)} \\
\midrule
Zero Shot (i)       & \textbf{0.2122} & 0.2359 & \textbf{0.2053} & \textbf{0.3378} & 0.3760 & \textbf{0.3210} \\
Non-cat.\ Prompt (ii) & \makecell{0.1298 \\ (-38.8\%)} & \makecell{\textbf{0.2973} \\ (+26.0\%)} 
& \makecell{0.1612 \\ (-21.5\%)} & \makecell{0.2056 \\ (-39.2\%)} 
& \makecell{\textbf{0.4822} \\ (+28.3\%)} & \makecell{0.2528 \\ (-21.2\%)} \\
Cat.-focused Prompt (iii) & \makecell{0.1972 \\ (-7.1\%)} & \makecell{0.2511 \\ (+6.4\%)} 
& \makecell{0.1904 \\ (-7.2\%)} & \makecell{0.3279 \\ (-2.9\%)} 
& \makecell{0.4119 \\ (+9.6\%)} & \makecell{0.3078 \\ (-4.1\%)} \\
\bottomrule
\end{tabular}
\end{adjustbox}
\end{subtable}

\caption{Evaluation of captions under two settings: (1) paragraph choice with original vs. filtered paragraphs, and (2) prompt strategies with zero-shot, non-category-focused, and category-focused prompts.}
\label{tab:prompt-opt}
\end{table}

\subsubsection{\textbf{Personalized Caption with Profile Figure} }

We observed that the selected captions achieved high ROUGE-1, ROUGE-2, and ROUGE-L recall scores (0.55, 0.30, and 0.47, respectively) but comparatively lower ROUGE-1, ROUGE-2, and ROUGE-L precision scores (0.40, 0.22, and 0.34, respectively) due to excessive irrelevant content. 
Few-shot prompting with the profile figures’ information improved BLEU scores (40–48\% gain) and precision-based ROUGE metrics (25–27\% gain), while maintaining competitive recall (5.60\% to 3.50\% loss), compared to the caption before refinement, indicating more concise and relevant captions without sacrificing essential content coverage.

\begin{table*}[t]
\centering
\scriptsize
\renewcommand{\arraystretch}{0.95}

\begin{subtable}{\linewidth}
\centering
\subcaption*{\textbf{(a) BLEU \& ROUGE-1 results}}
\begin{adjustbox}{max width=0.95\textwidth}
\begin{tabular}{lccccccc}
\toprule
\textbf{Condition} & B1 & B2 & B3 & B4 & R1-P & R1-R & R1-F \\
\midrule
All Generated Captions & \makecell{0.3008 \\ (+0.13\%)} & \makecell{0.2127 \\ (+0.66\%)} & \makecell{0.1621 \\ (+1.12\%)} & \makecell{0.1303 \\ (+1.56\%)} 
& \makecell{0.4217 \\ (+3.46\%)} & \makecell{0.5446 \\ (−2.37\%)} & \makecell{0.4103 \\ (+0.61\%)} \\
Selected Before Personalization & 0.3004 & 0.2113 & 0.1603 & 0.1283 & 0.4000 & 0.5500 & 0.4078 \\
Selected After Personalization & \makecell{\textbf{0.4226} \\ (+40.6\%)} & \makecell{\textbf{0.3093} \\ (+46.3\%)} & \makecell{\textbf{0.2377} \\ (+48.3\%)} & \makecell{\textbf{0.1825} \\ (+42.3\%)} 
& \makecell{\textbf{0.5090} \\ (+24.9\%)} & \makecell{\textbf{0.5266} \\ (−5.6\%)} & \makecell{\textbf{0.4972} \\ (+21.9\%)} \\
\bottomrule
\end{tabular}
\end{adjustbox}
\end{subtable}

\begin{subtable}{\linewidth}
\centering
\subcaption*{\textbf{(b) ROUGE-2 \& ROUGE-L results}}
\begin{adjustbox}{max width=0.95\textwidth}
\begin{tabular}{lcccccc}
\toprule
\textbf{Condition} & R2-P & R2-R & R2-F & RL-P & RL-R & RL-F \\
\midrule
All Generated Captions & \makecell{0.2335 \\ (+4.43\%)} & \makecell{0.3035 \\ (−1.75\%)} & \makecell{0.2285 \\ (+1.51\%)} 
& \makecell{0.3563 \\ (+4.26\%)} & \makecell{0.4655 \\ (−1.97\%)} & \makecell{0.3479 \\ (+1.28\%)} \\
Selected Before Personalization & 0.2200 & \textbf{0.3000} & 0.2251 & 0.3400 & \textbf{0.4700} & 0.3435 \\
Selected After Personalization & \makecell{\textbf{0.2858} \\ (+27.8\%)} & \makecell{0.2982 \\ (−3.5\%)} & \makecell{\textbf{0.2805} \\ (+24.6\%)} 
& \makecell{\textbf{0.4351} \\ (+27.3\%)} & \makecell{0.4499 \\ (−5.3\%)} & \makecell{0.4254 \\ (+23.9\%)} \\
\bottomrule
\end{tabular}
\end{adjustbox}
\end{subtable}

\caption{Comparison of caption performance (all, pre- vs.\ post-personalization); percentages show changes vs.\ baseline, with precision gains and recall losses.}

\label{tab:caption-performance-sub}
\end{table*}
\FloatBarrier 
\newpage
\bibliographystyle{colm2025_conference}

\clearpage
\appendix
\section{Appendix}
\begin{table*}[ht]
\centering
\renewcommand{\arraystretch}{0.95}

\begin{minipage}[t]{0.47\textwidth}
\begin{tcolorbox}[
  enhanced,
  colback=gray!8,
  colframe=black!60,
  fontupper=\scriptsize,
  title=\footnotesize Caption Ranking Prompt,
  boxsep=0.6pt,
  left=2pt,right=2pt,top=2pt,bottom=2pt,
  width=\textwidth
]
\scriptsize
\textbf{Context:}
\begin{itemize}[leftmargin=1em]
  \item Type: \{figure type\}
  \item Mention: \{mention\}
  \item Paragraph: \{paragraph\}
  \item OCR: \{OCR text\}
\end{itemize}

\textbf{Captions:} ``Cap. 1'', ``Cap. 2'', ``Cap. 3'' …

\textbf{Task:} Rank best$\to$worst by clarity, relevance, accuracy, tone.

\textbf{Output:} Ordered list + justifications.
\end{tcolorbox}
\end{minipage}
\hfill
\begin{minipage}[t]{0.47\textwidth}
\begin{tcolorbox}[
  enhanced,
  colback=gray!8,
  colframe=black!60,
  fontupper=\scriptsize,
  title=\footnotesize Caption Refinement Prompt,
  boxsep=0.6pt,
  left=2pt,right=2pt,top=2pt,bottom=2pt,
  width=\textwidth
]
\scriptsize
Refine caption style only; \textbf{no new info}.  

\textbf{Rules:}
\begin{itemize}[leftmargin=1em]
  \item Keep facts; no new entities
  \item Edit style/structure only
  \item Preserve IDs; concise
\end{itemize}

\textbf{Optional Demos:} Mention, Paragraph, Refined caption (×3)

\textbf{Target:}  
Mention: \{mention\_tgt\} \\
Paragraph: \{para\_tgt\} \\
Initial: \{init\_cap\} \\
Length: \{len\_hint\}  

\textbf{Output:} Final caption only.
\end{tcolorbox}
\end{minipage}

\caption{Prompt templates: (A) ranking and (B) refinement.}
\label{fig:caption-ranking-side}
\end{table*}
\end{document}